# Real Time Detection Free Tracking of Multiple Objects Via Equilibrium Optimizer

1st Djemai CHAREF-KHODJA*
*Electrical Engineering Department*
*Biskra University*
Biskra, Algeria
Email: djemai.charefkhodja@univ-biskra.dz

2nd Toumi ABIDA
*Electrical Engineering Department*
*Biskra University*
Biskra, Algeria
Email: a.toumi@univ-Biskra.dz

*Abstract*—**Multiple objects tracking (MOT) is a difficult task, as it usually requires special hardware and higher computation complexity. In this work, we present a new framework of MOT by using of equilibrium optimizer (EO) algorithm and reducing the resolution of the bounding boxes of the objects to solve such problems in the detection free framework. First, in the first frame the target objects are initialized and its size is computed, then its resolution is reduced if it is higher than a threshold, and then modeled by their kernel color histogram to establish a feature model. The Bhattacharya distances between the histogram of object models and other candidates are used as the fitness function to be optimized. Multiple agents are generated by EO, according to the number of the target objects to be tracked. EO algorithm is used because of its efficiency and lower computation cost compared to other algorithms in global optimization. Experimental results confirm that EO multi-object tracker achieves satisfying tracking results then other trackers.**

## I. INTRODUCTION

Multi-object tracking (MOT) is one of the most important and active field of search in the domain of vision by computer, because of its wide application areas. It is often used in surveillance systems [1], robotic vision, and other industrial applications such as surveillance video synopsis [2].

The task of Multi-object tracker is to estimate the position for every moving object and assign to it a unique ID.

However, MOT is a very challenging task due to situations such as: partial or full occluded objects, complex and moving background, or objects temporarily leaves the field of view, which can easily decrease an MO tracker performance.

There are two prominent approaches in MOT, the first one is detection-based tracking (DBT), and the second is detection-free tracking (DFT) [3]. Our framework is based on the second approach where the detection is done manually in the first frame, this approach is more popular, but it was given a very few attention compared to the first approach.

The second approach is based on a local search strategy by solving the problem of data association object-by-object and frame by frame. This class is similar to the probabilistic based trackers such as: Kalman filter (KF) [4], and particle filter (PF) [5], metaheuristique trackers [6-10], and deterministic trackers such as: Mean-Shift [11, 12] ...etc, used for single object tracking (SOT), and they have provided good performance. Thereafter many extensions for these trackers have been proposed to deal with multiple object-tracking frameworks. For mean-shift tracker for example, Gao et al. [13] proposed to improve mean-shift to be able to deal with MOT framework. Yuxin et al. in [14] and Mahmoud et al. in [15] proposed frameworks based on particle filter.

Particle Filter (PF) is the most used in MOT, however, in complex tracking environment, its tracking results are often affected, because its generated particles tend to move toward regions of high posterior probability, in particular, when two or more objects come close together or overlap.

Concerning metaheuristic MOT techniques, which are our related works, to the best of our knowledge, there have not been so much works on this line.

In the literature there is no specific definition of real time tracking [16, 17] either for SOT or MOT. Also there is no reference speed required for a tracker to be said real-time working. Instead, it is said that a model is real-time if it can give an output as fast as or faster than, the frame rate of the input video sequence.

And because that some applications require a certain tracking frame rate while others are not, we can assume that real-time tracking is depending on the application. For some applications such as robotics and surveillance, shorter delay and higher frame-rate might be necessary, and because of the challenges mentioned previously, MOT is considered to be a very hard task. In this case, we propose a way to attain real-time performance by reducing object features, which is the aim of the present work.

In this paper and in order to attain fast real time tracking we propose to reduce object resolution in order get very fast tracking.

Based on what was mentioned before, our main contributions can be summarized in these points:

- The first work to propose the recent Equilibrium optimizer (EO) for detection free MOT framework.

- Reduce the resolution of objects to reduce computation complexity.
- The use of Bhattachrya distance as fitness function in MOT.
- The performances of the proposed approach are assessed on both SOT and MOT.

The rest of the paper is organized into seven sections. In the next section we explain our vision and analysis of the related works. In section 3, we present in detail the Equilibrium Optimizer (EO). Section four, present kernel color weighted histogram used as object presentation for EO-SOT and EO-MOT. In the fifth section, we present the EO based tracker for MOT system. Section six we demonstrate the performance of EO tracker by giving a comprehensive comparison of the results with other reference trackers. Finally, section 7 gives the conclusion of this work.

## II. RELATED WORKS

Concerning MOT meta-trackers, to the best of our knowledge, there is no much work that has been done in this line, for example: M. Thida et al. [18] used particle swarm optimizer (PSO) as a searching approach, to find the most similar candidate, In their approach the MOT process was formulated as a repetitive optimization, (Not a parallel optimization). They have modeled the target objects by the covariance matrix of color feature, and thereafter added SIFT feature descriptor to tackle covariance color matrix lack of instability in background clutter. The fitness function is defined as generalized eigenvalues of covariance matrix; their approach provided good results in heavy occlusion, erratic motion, and illumination changes, without any comparison to other trackers.

Zhang et al. [19] proposed an approach based PSO, in which the interactions between the objects are modeled as species repulsion and competition, their approach is a set of individual trackers.

Hsu et al. [20] have presented a framework by hybridizing PSO with PF, with a switching mechanism. PSO is used in detection then it switches to PF to complete the tracking task of the objects. They have modeled the targets by their Gray-level histograms, and speeded up robust features (SURF) descriptor to handle the change in size of the objects.

Kwolek et al. [21] proposed MOT approach based on PSO. In order to improve its accuracy they have proposed a discriminative appearance model. The fitness function is based on the covariance matrix of multi-patch regions. However, according to their experience, PSO swarms size N, and iterations Iter should be "N=40 & Iter=300" to achieve its best results, which is enormous regardless of the implementation platform.

Hoang et al. [22] proposed pedestrian tracking using Bacterial Foraging Optimization (BFO) algorithm by segmenting the pedestrian's body into parts, which is demonstrated to provide better performance over traditional PSO and PF methods.

Guang Liu et al. presented an improvement to PSO named as weight adjusted particle swarm optimizer (WAPSO) [23], in order to solve the problem of pre-mature convergence and diversity loss of the traditional PSO, furthermore according to them, this improvement permits to tackle the problem of computational cost and occlusion in MOT. In their experiment they have proposed root sum squared errors (RSSE) as a fitness function, and their experimental results showed that their proposed algorithm was more accurate, and faster than categorized PSO, and traditional PSO.

Arababadi et al. [24] used Firefly algorithm (FFA) and feature similarity index (FSIM) as fitness function, FSIM is an algorithm that was introduced by zhang et al. [25], to assess the similarity of two images.

Most of these related works used particle swarm optimizer (PSO) for MOT, by tuning its parameters, or optimizing the features models of the target objects, and most of them used parallel running of the metaheuristic algorithm to gain speed of processing.

## III. EQUILIBRIUM OPTIMIZER (EO) ALGORITHM

Equilibrium Optimizer (EO) is a new metaheuristic algorithm, population-based, and physical inspired optimization algorithm proposed in 2019 by Afshin Faramarzi, et al. [26].

The algorithm mathematically mimics the adjustment of mass balance to appreciate both equilibrium and dynamique states; the inspiration of EO is a simple dynamic mass balance on a system of controlling volumes, in which the mass balance equation provides the underlying physics for the conservation of mass entering, leaving and generated.

In EO algorithm, each search agents randomly readjust their positions according to the best solutions obtained so far, to finally reach optimal result (equilibrium state). A mechanism called "**Generation rate**" is used to boost EO's ability in exploration, and exploitation to avoid local optimums.

### A. Agents Initialization and fitness evaluations

In EO, each search agents (position) mimics a concentration. In addition, like any other metaheuristic algorithms, EO starts with a random initial position (concentration) of its search agents as follows:

$$CT_i^{initial}=CT_{min}+R.(CT_{max}-CT_{min})_{\ i=1,2,...n} \qquad (1)$$

Where $CT_i^{initial}$ denotes the initial vector position (concentration) of the agent $i$, and, and $CT_{min}$ & $CT_{max}$ denotes respectively, the minimum and maximum border values of the search space, R is a random vector in [0, 1], and n is population size. Thereafter, their fitness function are evaluated and sorted to determine the balance candidates.

### B. Equilibrium pool

The global optimum is achieved when the equilibrium state is established.

$CT_{eq5}$ denotes the arithmetic mean of the best four agents ($CT_{eq1,...4}$) Identified during the whole optimization process these agents, helps in exploration ability, while $CT_{eq5}$ helps in exploitation ability.

These five agents named as equilibrium candidates are used to construct another vector named as the equilibrium pool.

$$CT_{eq,Pool}=\{CT_{eq1},CT_{eq2}, CT_{eq3}, CT_{eq4}, CT_{eq5}\} \qquad (2)$$

## C. Exponential term:

The exponential term (F) is used to update concentration; it helps EO to balance between exploration and exploitation.

$$F = a_1 \cdot \text{Sign}(R-0.5)*(\exp^{-\lambda \cdot t} - 1) \quad (3)$$

$$t = (1 - Iter / IterMax)^{(a2 \frac{Iter}{IterMax})} \quad (4)$$

Where:

**Iter** and **IterMax** denote respectively the current iteration, and the maximum number of iterations, since the changeover can vary over time in a real control volume, λ is assumed a random vector in [0, 1]. a1, a2 are constant values used respectively in conducting exploration and exploitation.

## D. Generation rate:

Improving the exploitation phase is carried by the generation rate, which aims to provide the best solution, expressed as follows:

$$Gr = Gr_0 \cdot F \quad (5)$$

where:

$$Gr_0 = GCP \cdot (C_{eq} - \lambda \cdot CT_i) \quad (6)$$

$$GCP = \begin{cases} 0.5 \, r_1, r_2 \geq GP \\ 0, r_2 < GP \end{cases} \quad (7)$$

Where $r_1$ and $r_2$ are random set in [0,1], and GCP is the vector that controls the Generation rate parameter. The term Generation probability (GP) determines how many search agents use the GCP parameter.

*GP* equals to 0.5 permits to achieve a good balance between exploration and exploitation. Finally, the rule for updating EO's search agents is as follows:

$$CT = CT_{eq} + (CT - CT_{eq}) \cdot F + \frac{Gr}{\lambda \cdot V}(1 - F) \quad (8)$$

Where:

F is already defined in Eq. (3), and V is constant equals to one.

## E. The memory saving of the search agents

EO algorithm has a similar mechanism of Pbest in PSO, in which, each search agent with its fitness value in the current iteration, update their values, if they were better than the previous ones.

This mechanism helps in the exploitation's ability, but it can raise the chance of falling into local optimum, if the algorithm does not have a good global exploration capability.

The pseudo code of EO algorithm is depicted in Fig.1

## IV. KERNEL COLOR WEIGHTED HISTOGRAM USED AS A FEATURE PRESENTATION (KCWH)

Selecting an object feature is a crucial step in any visual tracking system, and color feature is the most used object presentation in visual tracking. The principal reason is, despite the many changes that can occur to the object, the color is the most conserved feature of the object.

Before evaluating the object feature we need to evaluate the object size, which is defined by the following:

$$SzObj = \sqrt{W_i * H_i} \quad (9)$$

SzObj stands for the object's size, and Wi, Hi are respectively, the initial width and height of the bounding box that surround the object in the first frame.

The object size is reduced using **imresize** Matlab function. If the object size is superior or equal than of defined threshold 50, the object is resized to the scale 1/3 using **imresize** Matlab command, if not then no resize. Afterwards, the presentation called as: kernel color-weighted histogram (KCWH) [8,11] is carried to model the object.

Resizing the grounding box of the object template removes unneeded pixel information, and reduces computation complexity, because it reduce the size of the histograms to be computed and compared.

It should be noted here that, when resizing the object template the frame sequences are also resized, and the position is also multiplied by the same scale 1/3, to get the right position of the object.

```
1. Initialize the position of N search agents.
2. Assign values for the parameters v,a1,a2,GP.
3. While Iter < IterMax Do
4. For i=1 to N do
5. Evaluate fitness values of the i-th search agent
6. Update (CTeq1, CTeq2, CTeq3, CTeq4) and their corresponding Fitness
   according to B.
7. End For .
8. Set CTave as the average position of the 4 best search agent.
9. Set the equilibrium pool as in (2)
10. if (Iter >1) then establish memory saving.
11. Set t as in (4)
12. for j=1 to N do
13.    choose a random candidate from CTeq,Pool
14.    Generate λ & R.
15.    Set F as in (3)
16.    Set GCP vector as in (7)
17.    Set Gr0 as in (6)
18.    Set Gr as in (5)
19.    Update CT as in (8)
20. End for
21. Iter=Iter+1
22. End while
23. Return CTeq1
```

Fig.1 Pseudo code of EO

Generally, in visual tracking the object is defined inside a bounding box in the video frame. The kernel color weighted histogram of both objects candidates and templates are defined by:

$$Hg = C \sum_{i=1}^{n} k(\| x_i^* \|^2) \delta[b(x_i^*) - u] \quad (10)$$

Where:

- *k*: denotes the kernel that assigns big weights to the locations that are near the center of the target and vice versa.
- *C*: denotes normalization constant so that, $\sum_{u=1}^{m} Hg = 1$.
- **{X*_i}_{i=1...n}** : is the normalized pixel's locations of the target template centered at **0**.
- δ: is the Kronecker delta function.
- $b(x_i^*)$ are the quantized bins of **x*_i** for every distribution of color u.

Since the comparison is between histograms, Bhattacharyya distance (D) is the most used metric for comparison, which is defined by:

$$D[H1, H2] = 1 - \sum_{u=1}^{m} \sqrt{H_1(u)H_2(u)} \quad (11)$$

Where:

- m=16, is the bins' number in the histograms.
- $H_1$ and $H_2$ are the histograms of template and target being compared in the hue saturation value (HSV) color space.
- Fig.2 illustrates the flowchart of EO in SOT framework.

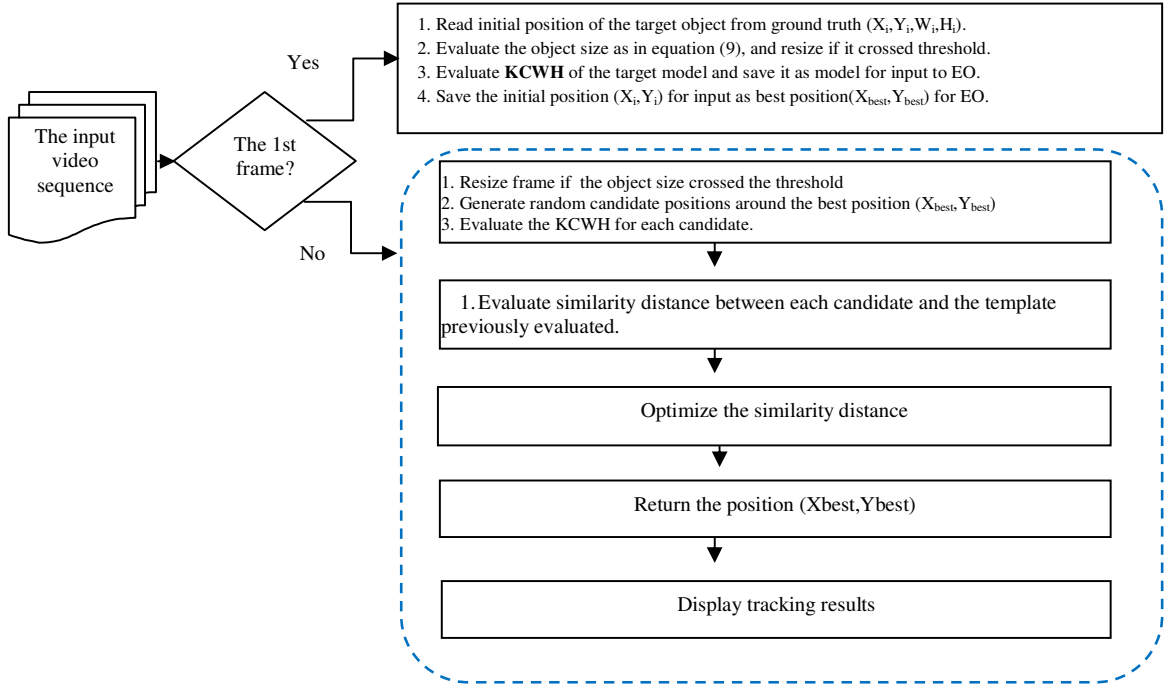

Fig.2 Proposed EO-SOT tracking algorithm

## V. EO TUNING PARAMETERS

In order to justify our chosen parameters, we need to study population size and iteration number effect on the performances of EO algorithm, for that raison we have chosen the Boy sequence as it contains the maximum number of challenges such as: fast motion, scale variation, blurred motion, in plane rotation and out of plane rotation.

The table 1 was given by running the algorithm 10 times and recording the average performance defined by average overlap or the area under curve as below:

$$Overlap = \frac{Area(B_I^p \cap B_I^g)}{Area(B_I^p \cup B_I^p)} \quad (12)$$

Where: $B_I^p$ and $B_I^g$ are respectively the predicted bounding box, and the ground-truth bounding box.

According to this experience, the combination agent numbers equals to 16 and maximum number of iteration equals to 4, seems to be good, as it permits good and stable results of EO-SOT algorithm. In the next experiments we will choose this combination to evaluate the performance of our proposed tracker.

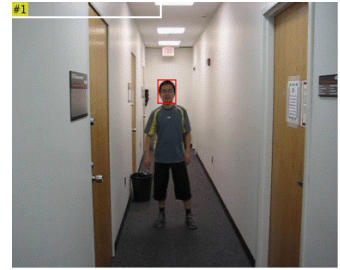

Fig.3 The first frame of the Boy sequence

## VI. EXPERIMENTAL VALIDATION

The proposed approach is validated on both SOT and MOT framework.

### VI.1. EXPERIMENTAL VALIDATION ON SOT

In this subsection we assess the performance of EO as a SOT quantitatively in precision and also in tracking speed.

### A. Precision evaluation

The precision is evaluated using 20 video sequences from the OTB2015 [27] and with comparison to other 5 reference trackers namely: adaptive structural local sparse appearance model (ASLA) [28], distribution fields for tracking (DFT) [29], locally orderless tracking (LOT) [30], kernel correlation filter (KCF) [31], robust visual tracking via multi-task sparse learning (MTT) [32]. The comparison is carried with EO for equilibrium optimizer without lowering resolution of objects, as a witness, and EO-LR with lowering resolution of objects to depict lowering resolution effect.

The usual way to compare between trackers in this benchmark is called as **one-pass evaluation metric (OPE)**, which is achieved by the initialization of every SOT from the ground truth position in the first frame and then display the success rate curves.

All our experiments are implemented in MATLAB R2013- on Intel I5-5200U 2.20 GHz CPU with 6 GB RAM.

In order to get the success plots namely: overlap plot Fig.3, and the precision plot Fig.4, we have used the toolkit available in website (www.visual-tracking.net). The reported curves are given respecting the protocol in [27] by using the same parameter values for all the sequences, and no manual tuning for every sequence. The legend of the location error curve contains a threshold scores at 20 pixels, while the legend of overlap curve contains area under-curve threshold at 0.5.

As displayed from both Fig.3 and Fig.4, both EO and EO-LR performs favorably against the other 5 state-of-the-art trackers. EO_LR achieved a precision rate of (53.8%) with a loss of only 1.8% compared to EO, and an overlap success rate of (46.1%), equal to EO.

It is also remarkable that the convergence curve of both EO and EO_LR based tracker are the best among the reference trackers, and this is due the powerful searching ability of EO combined with kernel color weighted histogram as feature

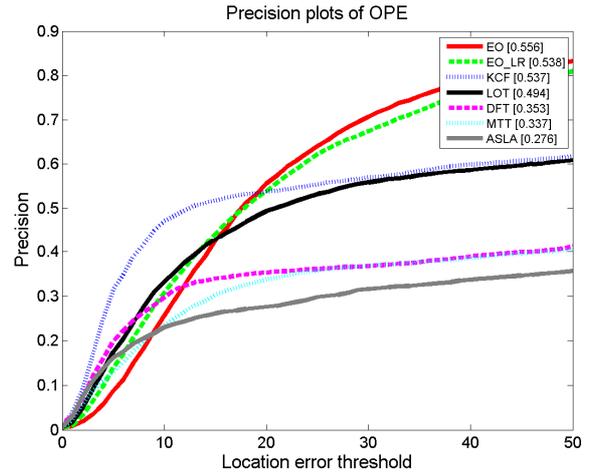

Fig.3 the comparison results of 7 trackers of the location error plot using OPE metric.

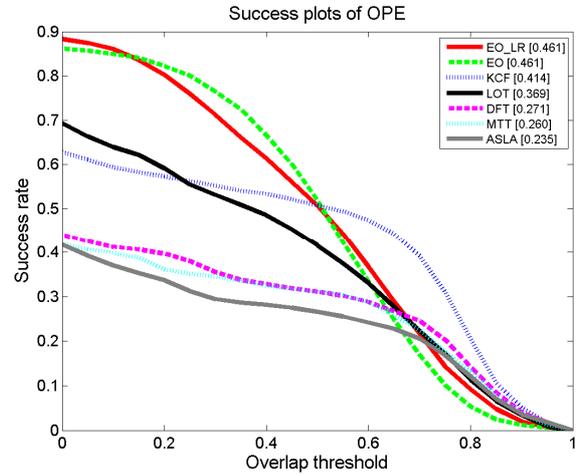

Table 1 the performance results of the average 10 independent running.

| | Pop size | 06 | 08 | 10 | 12 | 14 | 16 | 18 | 20 | 22 | 24 |
|---|---|---|---|---|---|---|---|---|---|---|---|
| Iter=01 | Mediane | 0.354 | 0.530 | 0.557 | 0.578 | 0.594 | 0.599 | 0.607 | 0.610 | 0.621 | 0.623 |
| | Max value | 0.485 | 0.561 | 0.566 | 0.587 | 0.605 | 0.613 | 0.618 | 0.629 | 0.629 | 0.640 |
| | Min value | 0.168 | 0.423 | 0.381 | 0.454 | 0.415 | 0.461 | 0.602 | 0.598 | 0.483 | 0.614 |
| Iter=02 | Mediane | 0.511 | 0.571 | 0.595 | 0.613 | 0.627 | 0.633 | 0.645 | 0.649 | 0.654 | 0.659 |
| | Max value | 0.555 | 0.585 | 0.606 | 0.623 | 0.635 | 0.648 | 0.655 | 0.653 | 0.664 | 0.667 |
| | Min value | 0.412 | 0.470 | 0.503 | 0.608 | 0.619 | 0.628 | 0.630 | 0.643 | 0.647 | 0.655 |
| Iter=03 | Mediane | 0.586 | 0.612 | 0.629 | 0.647 | **0.658** | 0.663 | 0.668 | 0.674 | 0.678 | 0.682 |
| | Max value | 0.597 | 0.625 | 0.641 | 0.654 | **0.667** | 0.671 | 0.675 | 0.685 | 0.685 | 0.690 |
| | Min value | 0.460 | 0.600 | 0.625 | 0.635 | **0.655** | 0.656 | 0.664 | 0.663 | 0.675 | 0.672 |
| Iter=04 | Mediane | 0.609 | 0.639 | 0.657 | 0.666 | 0.675 | 0.683 | 0.686 | 0.691 | 0.693 | 0.695 |
| | Max value | 0.622 | 0.643 | 0.666 | 0.674 | 0.684 | 0.688 | 0.694 | 0.695 | 0.697 | 0.699 |
| | Min value | 0.509 | 0.527 | 0.650 | 0.664 | 0.669 | 0.675 | 0.679 | 0.688 | 0.686 | 0.690 |
| Iter=05 | Mediane | 0.609 | 0.639 | 0.657 | 0.666 | 0.675 | 0.683 | 0.686 | 0.691 | 0.693 | 0.695 |
| | Max value | 0.622 | 0.643 | 0.666 | 0.674 | 0.684 | 0.688 | 0.694 | 0.695 | 0.697 | 0.699 |
| | Min value | 0.509 | 0.527 | 0.650 | 0.664 | 0.669 | 0.675 | 0.679 | 0.688 | 0.686 | 0.690 |
| Iter=06 | Mediane | 0.642 | 0.663 | 0.676 | 0.682 | 0.691 | 0.695 | 0.699 | 0.700 | 0.704 | 0.704 |
| | Max value | 0.650 | 0.666 | 0.682 | 0.687 | 0.695 | 0.699 | 0.702 | 0.704 | 0.707 | 0.707 |
| | Min value | 0.526 | 0.652 | 0.664 | 0.679 | 0.689 | 0.692 | 0.694 | 0.698 | 0.699 | 0.701 |

The bold values indicate the chosen combination performance

## B. Tracking speed

We assess the performance of EO-SOT using speed metric, in which we measure the FPS. Regardless of implementation platform, elements such as the size of the bounding box (BB) that surrounds the object, the feature used to model the object, the number of particles, and the number of iterations used in a metaheuristic algorithm; those are all parameters that affect the tracking speed.

In both the initialization and updating process in a metaheuristic algorithm, the speed of tracking measures the calculation's complexity, which is directly related to the number of objective function calculations. The number of fitness function assessments in EO-SOT was 42 in the earlier of our experiments.

The computation of FPS is given using **tic-toc** Matlab command, and because it gives us different values at each run, we collected in table 2 the average ten runs of FPS for three reference trackers to get good values for comparison.

Concerning tracking speed, KCF and KMS are our benchmark trackers, because they are well known for their use in real time applications as well as their higher tracking speed.

From left to right, the objects are sorted by their size, from smallest to largest. According to the table 2, EO tracker tracking speed decreases as the object size increases, which is not the case for KMS [11] and KCF [31], because they have different tracking mechanism.

KMS for example, is an algorithm that implements conditioned iterations, which why its speed have no direct relationship with the object's size.

EO and EO_LR for Girl and Boy sequences have the same performance because their size in equation (9) has not crossed the threshold.

In the Skating2.1 sequence the size crossed the threshold, but the tracking speed has not improved so much, because of the use of imresize Matlab function which increases a little the computation complexity.

In both the Skating2.1 and BlurBody sequences, we can see that our proposed tracker EO_LR has surpassed the tracking speed of both KMS and KCF.

## VI.2 EXPERIMENTAL VALIDATION ON MOT

Before starting experimental validation of EO on MOT, in this subsection we illustrate how to implement detection free EO-MOT.

The task of multi-object localization in every frame is considered as a parallel processing optimization problem.

In the following, we introduce our approach, which is used in our experimental studies.

We implement multiple agents for MOT according to the number of object to be tracked. First, we select multiple target objects, either manually from the screen or from the ground truth, then secondly we establish the feature model for all objects, then for each agent we associate an object model. The agents then process the target search in parallel processing to make the search quick, as every agent concentrate on its own target separately.

Fig.5 shows the EO-MOT parallel implementation in the case of k fitness functions.

Instead of a single population, several sub-populations are used, each of which has its associated fitness function. To implement EO based MO tracker, every target is represented by a bounding box ($BB_i$) defined by a vector $P_i = (x_i, y_i)$ where $\{x_i, y_i\}$ denotes its 2-D translation coordinate, under assumption that its width and height do not change considerably in time. EO algorithm searching area is defined by its lower boundary $LB_i$, and its upper boundary $UB_i$ defined by:

- $LB_i = C_i(t) - \min(W_i, H_i)$ (11)
- $UB_i = C_i(t) + \min(W_i, H_i)$

Where $C_i$, denotes the best location of the object $i$, found so far, set manually at each call as an agent position, and $W_i$ is the initial width of the BB of the *i-th* object, set from the ground truth. Then, every object is modeled by its kernel color histogram.

Table 2 comparison of tracking (FPS) of four different SOTs

|  | Girl | Boy | Jogging-1 | Skating2.1 | BlurBody |
|---|---|---|---|---|---|
| Object size (width*height) | (31*45) | (35*42) | (25*101) | (64*236) | (87*319) |
| EO | 33,59 | 17,72 | 14,42 | 12,15 | 5,71 |
| EO_LR | 33,59 | 17,72 | 27,08 | 17,11 | **16,96** |
| KMS | 26,10 | 17,52 | 20,47 | 9,34 | 9,98 |
| KCF | **56,83** | **27,79** | **34,47** | **20,41** | 15,92 |

The bold values indicate the best values

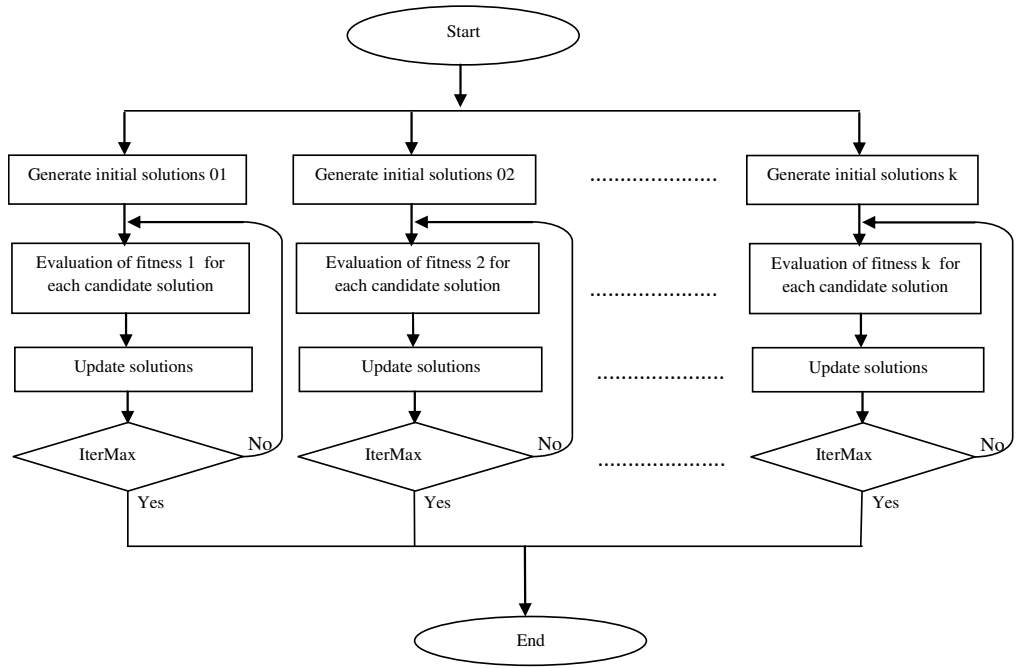

Fig.5 The flowchart of EO with a parallel implementation

To the best of our knowledge, the proposed MOT in the detection free framework that can be found in the literature, are scarce and their source codes are not available on the contrary to SOT, which is why we could not bring them for comparison.

In the following, we demonstrate the effectiveness of the proposed approach, and evaluate EO-MOT performance qualitatively and quantitatively using two video sequences from OTB2015 Benchmark [27]. The sequences names are '**Jogging**', '**Skating2-[1,2]**', in Fig.6, and they are freely available from the website: www.visual-tracking.net.

WAPSO proposed by Guang Liu et al. [23], FFA [24], and Harris hawks optimizer (HHO) proposed by Djemai et al. [9], serves as a benchmark to evaluate our proposed approach. For a fair assessment of the proposed approaches, we have used their proposed fitness function, the same initialization, the same population size, along with their recommended sensitivity's parameters, which are depicted in table 3.

Table 3 simulation parameters of every tracking algorithm

| Algorithm | Parameter | Value |
|---|---|---|
| **EO** | N : Population size | 14 |
| | Iter: Iteration number | 3 |
| | $a_1$ | 2 |
| | $a_2$ | 1 |
| | GP | 0.5 |
| | V | 1 |
| | Fitness function | KCWH |
| **WAPSO** | N : Population size | 14 |
| | Iter: Iteration number | 3 |
| | Wi | 1 |
| | C1 | 2 |
| | C2 | 2 |
| | $\alpha, \beta$ | 0.5, |
| | Fitness function | RSSE same as in [23] |
| **FFA** | N : Population size | 10 |
| | Iter: Iteration number | 3 |
| | $\beta_0, \gamma$ | 1 |
| | $\alpha$ | 0.5 |
| | Fitness function | FSIM same as in [24] |
| **HHO** | N : Population size | 10 |
| | Iter: Iteration number | 3 |
| | Fitness function | KCWH same as in [9] |

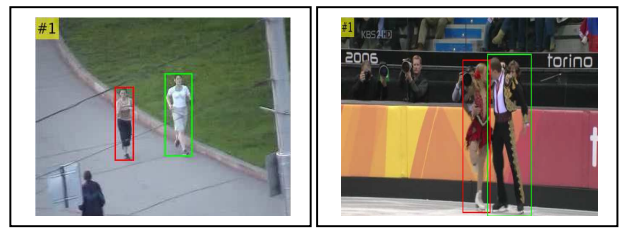

Fig.6 The first frames of the two sequences

namely from left to right: Jogging, Skating2.

The first video named as **Jogging,** Fig.6, on the left; two women doing a jogging, in this sequence EO-MOT performance is assessed under three challenging situations such as full occlusion, deformation, and out of plane rotations, which are real world situations.

The second video named as **Skating2,** on the right, the challenge is track a couple of man and woman skating, this sequence is more difficult than the previous, because of the challenging environment such as important Scale Variation, full occlusion, deformation, fast motion, and Out of plane rotation. The first object (Obj.1) is surrounded by a red grounding box, while the second (Obj.2) by a green one.

*A. Quantitative comparison*

In this section, we report three quantitative evaluations of the proposed MOT, namely: accuracy metric, precision metric, and also the speed of tracking metric [33].

*A.1 Precision metric*: it expresses how well accurate positions of objects are predicted[33].

Table 4 reports the average performance of ten independent runs. The average Intersection over union (IoU) Fig.7, and the average Euclidean distance (E-D) of the mentioned video sequences, and also the difference max-min values to illustrate the stability of the algorithm.

In this table EO-LR provided the best average performance of Euclidean distance with 26.31 pixels, and also the best IoU with 51%, along with EO and HHO.

In general, and in terms of precision EO and EO_LR provided good performance, because of two main reasons namely: the use of KCWH permits good presentation of the object model than RSSE and FSIM combined respectively with WAPSO and FFA, secondly EO and EO_LR provide good exploration of the search space, which is why they provided better average Euclidean distance than HHO.

Table 4, depicts the average performances of 10 runs of four reference trackers, while EO stands for equilibrium optimizer without lowering resolution of objects, and EO-LR for equilibrium optimizer with lowering resolution of objects.

*A.2 Accuracy metric:* this metric shows how many errors the tracker made in terms of missed objects, false positives, mismatches, and so forth [33], it also depicts how many mistakes made by the tracker concerning, misses, false positives, mismatches, ...etc.

In our case because we are in DFT framework, where the number of objects is already known in the first frame, we define the false negative rate ($FN_{rate}$) as follows:

$$FN_{rate}=100*(\sum N_t - \sum TP_t)/ \sum N_t \qquad (12)$$

Where $N_t$ is how many times the object is appeared in the frame t, TP is the number of true positive tracked object, and like in Fig.7, is observed when IoU prediction score exceeds the predefined threshold, generally 0.5. Table 5 depicts the average $FN_{rate}$ metric values of ten runs.

According to table 5, EO MOT provided the best lowest false negative rate, and then HHO provided the second best value. We can observe that EO_LR degraded its performance by 3.5%, which is acceptable compared to the other reference MOT.

Table 4 the average performance IoU and E-D of ten run of five MOT.

| | | | EO | | EO_LR | | WAPSO | | FFA | | HHO | |
|---|---|---|---|---|---|---|---|---|---|---|---|---|
| | | | Average IoU | Average E-D | Average IoU | Average E-D | Average IoU | Average E-D | Average IoU | Average E-D | Average IoU | Average E-D |
| Jogging | Obj. 1 | Average value | **0,59** | **10,18** | 0,54 | 14,63 | 0,17 | 87,12 | 0,56 | 16,11 | **0,59** | *10,64* |
| | | Max-Min value | **0,01** | 0,85 | **0,01** | 1,59 | **0,01** | 4,99 | 0,21 | 19,66 | *0,02* | 1,00 |
| | Obj. 2 | Average value | 0,56 | *18,14* | 0,59 | **17,72** | 0,49 | 45,48 | 0,48 | 26,89 | 0,56 | 18,46 |
| | | Max-Min value | **0,03** | *1,15* | **0,03** | 1,92 | 0,54 | 91,05 | 0,45 | 50,79 | **0,03** | 1,47 |
| Skating2 | Obj. 1 | Average value | *0,49* | 27,37 | 0,48 | 29,53 | 0,21 | 106,84 | 0,20 | 116,98 | **0,50** | **26,79** |
| | | Max-Min value | **0,01** | 1,24 | **0,01** | 0,69 | 0,08 | 43,20 | 0,25 | 111,26 | **0,01** | 1,69 |
| | Obj. 2 | Average value | 0,40 | 51,90 | **0,42** | **43,37** | 0,25 | 63,11 | 0,16 | 130,74 | 0,39 | 59,86 |
| | | Max-Min value | *0,03* | *14,11* | **0,02** | **6,13** | 0,53 | 101,86 | 0,08 | 46,90 | 0,04 | 21,04 |
| Average | | | **0,51** | *26,89* | **0,51** | **26,31** | 0,28 | 75,64 | 0,35 | 72,68 | **0,51** | 28,94 |

The bold values indicate the best performance, while the italic bold values indicate the second best performance.

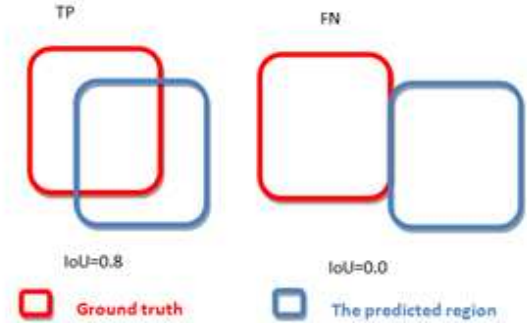
Fig.7 Object IoU

A.3 Tracking speed metric: table 6 reports the five trackers performance in terms of computation complexity by computing the tracking speed in frame per second (FPS). It can be seen from this table that EO-LR has the best tracking speed.

EO achieves better tracking speed results compared to HHO because it compute only 42 fitness values per object, while HHO compute 87 fitness function evaluations which justifies the average time cost between the trackers.

In terms of tracking speed EO_LR provided the best performance, because the size of the object is reduced considerably.

EO, EO_LR and HHO tracking speed are better compared to WAPSO and FFA, as they use respectively RSSE and FSIM which computation complexity are very high compared to KCWH, then according to our experience they are not suitable to be used in real time.

Table 5 the average FN metric values of ten runs.

|  |  | MOT | EO | EO_LR | WAPSO | FFA | HHO |
|---|---|---|---|---|---|---|---|
| Jogging | Obj. 1 | Average value | 32,44 | 46,84 | 75,18 | **25,37** | 34,69 |
| Jogging | Obj. 2 | Average value | 28,11 | **25,11** | 38,57 | 52,28 | 27,33 |
| Skating2 | Obj. 1 | Average value | **52,94** | 56,55 | 85,98 | 77,27 | 53,07 |
| Skating2 | Obj. 2 | Average value | 68,27 | **67,25** | 83,68 | 86,87 | 68,27 |
| Average |  |  | **45,44** | 48,94 | 70,85 | 60,45 | 45,84 |

The bold values indicate the best performance

Table 6 the average FPS values of ten runs.

|  |  | EO | EO_LR | WAPSO | FFA | HHO |
|---|---|---|---|---|---|---|
| Jogging | Average value | 13,11 | **14,76** | 4,31 | 0,40 | 10,55 |
| Skating2 | Average value | 8,52 | **15,61** | 1,89 | 0,10 | 5,06 |
| Average |  | 10,82 | **15,19** | 3,10 | 0,25 | 7,81 |

The bold values indicate the best performance

B. QUALITATIVE COMPARISON

In order to illustrate qualitatively the tracking performance of the proposed tracker with the three reference MOT WAPSO and FFA, we reported in table 7 and table 8 snapshots of the two mentioned sequences, in different challenging situations.

In table 7, the Jogging sequence, and in frame #46, EO_LR and FFA got the accurate position of the obj.2 while WAPSO lost the object. In frame #65, obj.1 was half occluded and EO_LOR got the exact position, FFA lost a part of the object, and a tracking failure of WAPSO. In frames #81 and #258, the two objects undergone deformations and change in size, only EO_LR could get the accurate targets' position, while the others failed.

In table 8, the Skating2 sequence. In frame #14, only FFA failed to get the exact position of Obj.2 because it undergone partial occlusion and out of plane rotation. In the frame #25, WAPSO lost Obj.2, and FFA lost both Obj.1 and Obj.2 because it undergone deformation and out of plane rotation. In frame #90, WAPSO lost Obj.2 and FFA lost Obj.1 and only EO_LR managed to track them. In frame #104, Obj.1 was almost fully occluded behind Obj.2 results the failure of all the trackers except EO_LR.

VII. CONCLUSION

In this paper, we have implemented EO algorithm in multi-object tracking framework in lots of challenging situations, by proposing a new architecture of MOT using EO by reducing the object feature for less computation cost, we also proposed to tune population size and iterations number according to the video frame rate to attain real time tracking.

To demonstrate the tracking performance of the proposed approach, quantitative and qualitative comparisons on both SOT and MOT proved that EO provided high real time speed of tracking and good accuracy.

To the best of our knowledge, this is the first time for EO to be used in SOT and MOT framework, and our initial experimental results show that EO has provided satisfying results in accuracy and speed. Due to its tracking speed, EO_LR can successfully be applied in real-time MOT. Future works can be dedicated to the use of the EO-MOT in the detection based framework, in a new promising application such as traffic monitoring of public pedestrian, to verify the respect of the safety distance between people, and the verification of wearing a mask to slow the spread of COVID-19.

Table 7. Tracking results in Jogging sequence, object sizes: (25x101)+(37x114).

| Frame N° challenge | EO_LR | WAPSO | FFA |
|---|---|---|---|
| 46 Deformation + Half occlusion |  |  |  |
| 65 half occlusion |  |  |  |
| 81 Deformation + Half occlusion |  |  |  |

| Frame N° challenge | EO_LR | WAPSO | FFA |
|---|---|---|---|
| 258 Deformation + Change in size | | 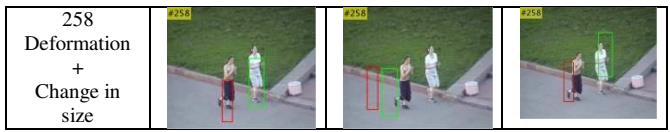 | |
| | | | |

Table 8 tracking results in Skating2 sequence object sizes: (64x236)+(103x251).

| Frame N° challenge | EO_LR | WAPSO | FFA |
|---|---|---|---|
| 14 half occlusion + Out of plane rotation | | | |
| 25 Deformation +Out of plane rotation + Half occlusion | | | |
| 90 Deformation +Out of plane rotation | | | |
| 104 Partial occlusion+ Out of plane rotation+ deformation | | | |